\documentclass[conference]{IEEEtran}
\IEEEoverridecommandlockouts
\usepackage{cite}
\usepackage{algorithmic}
\usepackage{amsmath,amssymb,amsfonts}

\usepackage{verbatim}
\usepackage{graphicx}
\usepackage{textcomp}
\usepackage{xcolor}
\usepackage{hyperref}
\def\BibTeX{{\rm B\kern-.05em{\sc i\kern-.025em b}\kern-.08em
    T\kern-.1667em\lower.7ex\hbox{E}\kern-.125emX}}
\begin{document}

\title{Ontological Component-based Description of Robot Capabilities\\
\thanks{This work has been supported by the Effective Learning of Social Affordances (ELSA) project ANR-21-CE33-0019 and by the MUGERI project via the Défi Clé Robotique Centré sur l'Humain de la région Occitanie.}
}


\author{Bastien Dussard$^{1}$ \and Guillaume Sarthou$^{2}$ \and  Aur\'elie Clodic$^{1}$ \\ 
{
}\\

\thanks{$^{1}$LAAS-CNRS, Universit\'e de Toulouse, CNRS, Toulouse, France {\tt\small firstname.surname@laas.fr}}
\thanks{$^{2}$IRIT, Universit\'e de Toulouse, CNRS, Toulouse, France {\tt\small firstname.surname@irit.fr}}}

\maketitle

\begin{abstract}

A key aspect of a robot's knowledge base is self-awareness about what it is capable of doing. It allows to define which tasks it can be assigned to and which it cannot.
We will refer to this knowledge as the Capability concept. As capabilities stems from the components the robot owns, they can be linked together. In this work, we hypothesize that this concept can be inferred from the components rather than merely linked to them. Therefore, we introduce an ontological means of inferring the agent's capabilities based on the components it owns as well as low-level capabilities. This inference allows the agent to acknowledge what it is able to do in a responsive way and it is generalizable to external entities the agent can carry for example.
To initiate an action, the robot needs to link its capabilities with external entities. To do so, it needs to infer affordance relations from its capabilities as well as the external entity's dispositions.
This work is part of a broader effort to integrate social affordances into a Human-Robot collaboration context and is an extension of an already existing ontology.

\end{abstract}


\section{Introduction}

In order to have an efficient collaboration between robots and humans, self-awareness of what each agent can do in a given context to fulfill the goal is a key step.

Indeed, the tasks an agent is able to perform highly depend on the its capabilities. In the context of collaboration, tasks can be decomposed into a sequence of sub-tasks, each assigned to one of the agents. Such attribution would be based on the capabilities of the involved agents.
As some sub-tasks of the plan to complete the goal can be better suited for a specific agent, an efficient collaboration between agents relies on knowledge of what the agents are capable of doing.
Therefore, modelling what an agent is capable of doing is necessary, either in terms of how it can act on the environment (e.g grasping objects) or what it can perceive in the environment (e.g detecting human presence). This modelling can leverage the components owned by the agent as they are the reason behind its ability to do something.
Collaborative tasks often rely on object manipulation such as assembling parts in the industry or cleaning the table in a home-helper context. The presence of other agents in the process requires to also take into account other modalities such as communication or perception.
Nevertheless, having a capability is not a sufficient condition to apply it to any object in the environment. Some objects provide action possibilities, i.e dispositions, which can be performed only if the agent has the corresponding capabilities as well as matching other constraints.


This intuitive idea has been formalized by Gibson with the concept of affordance. In~\cite{gibson2014ecological}, he defines affordance as \textit{"what it [the environment] offers the animal, what it provides or furnishes, either for good or ill"}. In addition, he states that these affordances exist in relation to the agent's capabilities. This concept has been built upon through the years through multiple works. Notably, Norman in \cite{norman2002design} defines affordance as \textit{"(perceived) possibility for action"}, stating that in order for an affordance to exist it must be directly perceptible by the agent. This concept has been of interest to the robotics community for a few years now and has been mostly applied to manipulation tasks and action prediction.
Several affordance formalisms have been proposed over the years in the ecological psychology domain \cite{turvey1992affordances, chemero2003outline, stoffregen2003affordances, csahin2007afford, steedman2002formalizing} and some proposals adapting those theoretical works into the robotics field have been developed \cite{montesano2007affordances, cruz2016training, barck2009learning, bessler2020formal}.


Even if each formalism represents a different point of view, they all link an agent (the actor) to an entity (the object). These formalisms are sufficient in robotics for object manipulation but in the context of human-robot interaction (HRI), the presence of other agents brings new possibilities. The latter are commonly referred as \textit{social affordances}. In \cite{carvalho2020social}, Carvalho defines them as \textit{"possibilities for social interaction or possibilities for action that are shaped by social practices and norms"}. In more straightforward words, we can define them as \textit{action possibilities offered by the presence of a set of social agents}.


In the robotics community, describing a basic action a robot can do is either referred to as \textit{Capability} or as \textit{Skill}, but these terms have different meanings.
\textit{Capability} is defined as \textit{The power or the ability to do something} in \cite{olivares2020review}. Hence, a capability can either be an action which has a direct impact on the environment (e.g grasping an object) or no direct impact (e.g detecting human presence). \textit{Skill}, on the other hand, refers to \textit{the ability to do something well}~\cite{olivares2020review}. Therefore, if we refer to a robot having a \textit{Skill}, it needs to have the required capabilities to do so. As this work focuses on the ability of an agent to perform an action regardless of the execution, we will only refer to agents' capabilities. Having a standardized way of representing the \textit{Capability} concept, following the standardization effort in \cite{prestes2013towards}, would benefit the robotics community by easing the integration process of external ontologies.


This work falls within the DACOBOT~\cite{sarthou2021director} architecture which is a knowledge-centered robotic architecture. It is based on an ontological description of the robot's knowledge in which the elements of the environment and some simple robot descriptions are present. Nevertheless, as we saw above, to allow a robot to reason about the actions it could perform regarding the elements of the environment, we first have to provide it with the knowledge of its capabilities.


In this paper, we present an ontological component-based modelling of capabilities towards an HRI context.
The main contribution of this work is an \textbf{inference mechanism over agent's capabilities} based on the components it owns and a way of linking those two concepts.


\section{Related Work}


Multiple approaches tackled the problem of representing the capabilities of robotic agents in an ontology and relating them to the components available on the robot. Most approaches \cite{umbrico2020ontology, beetz2018know, diab2019pmk, stenmark2015knowledge} use this link in an "action-oriented" manner, by having actions at a starting point, and verifying via queries if those actions are do-able (meaning if the robot has the right set of components/capabilities). 
On the opposite, our approach focuses on a "component-oriented" description of the capabilities. It allows for a more generic way of representing the capabilities of an agent while making autonomous the process of verifying them.

\subsection{Components Description}

The \textit{Component} concept refers to hardware and software parts which the robot is either composed of, or uses in its processes (e.g gripper, camera, pose estimation algorithm). The description of these devices allows to represent the body parts of the robot as well as how they are connected.

Robot hardware components are finely detailed in \cite{beetz2018know}, as it is based on Semantic Robot Description Language (SRDL) which automatically imports the hardware components via the robot Universal Robot Description Format (URDF) file, and some software components are included in the \textit{DesignedSoftware} class (e.g LocalizationSoftware, MappingSoftware or PerceptionSoftware).
While the hardware components description of this work is more precise than what is needed for our purpose, the software components part doesn't have the granularity required for the HRI perception's purpose. On the other hand, the other approaches have a very straightforward description of the robot because their robots are robotic arms and not more "human-shaped" robots. \cite{umbrico2020ontology} has an abstract description of the sensing components as they use and extend Semantic Sensor Network Ontology (SSN), which is an ontology for describing sensors and their observations. Its hardware component part is also straightforward as it features robots with simple configurations. It is therefore too elementary for reasoning about hardware-linked capabilities. In \cite{diab2019pmk}, robots are described by hardware sub-parts (like gripper or mobile base). Software components (like GeometricPlanningAlgo or LocateObjAlgo) and the link between hardware and software components is done via a \textit{has\textunderscore algorithm} property. This hardware components description better suits our purpose, but also lacks details on the software components part as it was used for manipulation and not for interaction.
\cite{stenmark2015knowledge} uses a \textit{Device} class which includes robot configurations as well as sensor components or 
grippers. As the three latter's field of application is a manufacturing context, the robot components involved - either software or hardware - do not represent robots which are more designed towards interaction (e.g human detection-related components or communication components).

\subsection{Capabilities Description}

The \textit{Capability} concept refers to a primitive action an agent can do. This concept is tightly related to the \textit{Component} concept in our work as it is the reason capabilities are available to an agent.


The \textit{Capability} concept is included in some approaches with this terminology \cite{beetz2018know, umbrico2020ontology} or referred to as the \textit{Skill} concept \cite{stenmark2015knowledge}. As the applications and goals between those works are different, the granularity over the capability description differs.
Indeed, \cite{stenmark2015knowledge} provides a manufacturing skills description which includes elementary capabilities such as opening the gripper, but also compound ones such as displace an object or sensor-related processes. \cite{umbrico2020ontology} also provides a manufacturing capability description which details basic acting capabilities and higher-level ones, but uses and extends SSN for the perception description to link the properties of objects and the capabilities of available devices.
On the other hand, \cite{beetz2018know} provides a task-agnostic capability description but may lack a detailed perception capabilities description for our use-case. Those last two articles use dependency conditions to connect capabilities to components or low-level capabilities and this kind of feature matches our work's goal.


\subsection{Reasoning about capabilities and components}

Most of the methods use Prolog predicates to verify the feasibility of an action, which translate to ensuring that the robot has the right set of components and capabilities. \cite{beetz2018know} uses a Prolog predicate with the \textit{dependsOnCapability} and \textit{dependsOnComponent} object properties to reason about which capabilities are available to the agent. \cite{umbrico2020ontology} uses Semantic Web Rule Language (SWRL) rules as the main mechanism to infer informations about capabilities and components. It uses the rule-inferred \textit{canPerform} object property linking an agent, an action, a capability and the requirements for this action. In this property, the \textit{hasCapability} property is not inferred based on the agent components but asserted for each component. It is then rule-inferred to link the agent owning the component to this capability. Another reasoning element of this work is about inferring if a robot component is an acting or a sensing part based on the capability it is used in (either acting or sensing capability). Therefore the component types are only referred to via this dichotomy and not in more detailed categories. \cite{diab2019pmk} uses the predicate \textit{feasible(Artifact, ?Action)} to verify that the robot has the capability to execute actions but without reasoning about the components required for it. \cite{stenmark2015knowledge} doesn’t reason about capabilities and components, but only ensures for each skill that the necessary components are available.

To summarize, in most of the approaches the relations between capabilities and components are hard-coded and the capabilities are focused on a few tasks rather than describing more general ones. 
Moreover, as most methods use queries to verify the capabilities of the agent, the process isn't dynamic and there is no inference over the capabilities based on the components they own.

\section{Ontology description}

Our work takes place in a broader effort to unify knowledge representation in robotic architecture. Indeed, having a unique and centralized knowledge base in a robotic architecture prevents knowledge duplicate and thus possible inconsistencies. It also allows more powerful reasoning as it takes into account the entire knowledge. The ontology representation has been chosen for the robot knowledge base due to its high expressiveness and the inferences mechanisms supported by First-order logic (FOL) and Description logic (DL). In this section, we first present our agent description and the capabilities representation before introducing how those capabilities can be inferred as being owned by an agent based on its components. The objective of this knowledge base extension is to provide \textbf{self-awareness knowledge} about what the robot can do given its components and the environment in a responsive way.

\subsection{Agent description}

Our agent description is oriented toward the components composing the robot. In such a way, we want to link our agent both to its body parts, its actuators, its sensors, and its software components. As illustrated in Fig.~\ref{fig:localisationCapability}, the simplest description would be to use a \textit{hasComponent} property. Within our example, the \textit{pepper} robot is composed of a head body part, itself equipped with a realsense camera, and an object tracking algorithm. Such description is close to the actual robot composition and could be automatically generated based on a URDF as in \cite{beetz2018know}. For a more precise semantic description, one could refine the \textit{hasComponent} property to distinguish between actuators, sensors, and software components.

Nevertheless, for the following inferences, we need all the components to be directly linked to the robot entity. To do so, we chose to make the \textit{hasComponent} property transitive. As shown in Fig.~\ref{fig:localisationCapability}, this axiom allows the realsense individual to be inferred as a component of our robot, represented by the \textit{pepper hasComponent realsense} relation.


\subsection{Capability hierarchical view}


We introduce a meta-class \textit{Capability}, divided into sub-classes corresponding to "fields of capability". Each of those fields has subclasses representing lower-level capabilities which are organized in a semantic way. For example, the \textit{Communication} capability field has subclasses for several means of communicating information, which include displaying a facial expression or pointing towards something.



We consider that a capability is enabled if an agent has the right set of components. To represent it, we chose to represent capabilities with class equivalence based on the required components. For example, the \textit{ObjectLocalisationCapa} capability is stated to require an agent having a \textit{Camera} component and an \textit{ObjectTracker} component. Therefore we can represent this capability as \textbf{\textit{ObjectLocalisationCapa} being equivalent to having \textit{some ObjectTracker} and \textit{some Camera components}}.


While some capabilities only require components to be enabled, others also require capabilities to exist. 
For example, pointing towards something requires a component to point with, such as a hand, but also the capability to localize the object to be pointed at. The only difference between component and capability dependency conditions is that the latter is referred to by its class and not by a \textit{hasComponent} property. In such a way, \textbf{\textit{HandPointingCapa} would be equivalent to \textit{ObjectLocalisationCapa} and \textit{hasComponent} some \textit{Hand}}.



\subsection{Component-based inference over capabilities}

\begin{figure}[t!]
\centering
\includegraphics[width=1.0\linewidth]{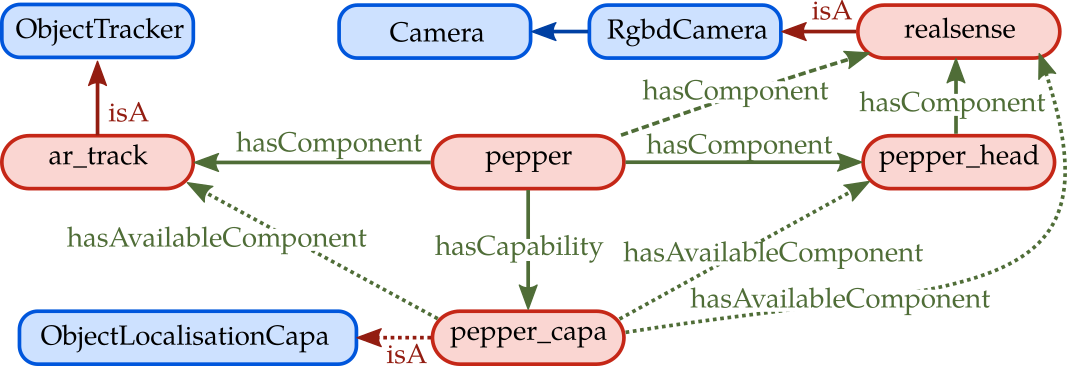}
\caption{\label{fig:localisationCapability} A graphical representation of an ontology describing a robot with its components and its inferred capability. The blue blocks correspond to classes and the red ones to individuals. Red arrows represent inheritance relations, with the dashed one representing inferred relation by equivalence. The green arrows relate to properties, with the dotted and dashed ones corresponding to inferred relations (respectively by transitivity and by chain axiom).}
\end{figure}

In the current litterature as well as our work, the \textit{Capability} concept is divided into several classes. Thus if one wanted to represent the capabilities of an agent, it would require the creation of an individual for each capability offered by combinations of components and lower level-capabilities. 

On top of requiring knowledge engineering to define beforehand all the possible capabilities of an agent in terms of ontology individuals, this usage is not convenient for a dynamical process as we cannot dynamically create individuals. Our approach rather choses to represent the capabilities of an agent with only one instance linked to the agent via the \textit{hasCapability} property. We are aware that such representation can lack semantic precision but it allows to avoid having to instantiate each capability in single individuals.
Regarding this single capability instance, our goal is to make this individual inherit from the capability classes "enabled" by the components owned by the agent. Thus, if we used the \textit{hasComponent} relation to define the equivalence relations representing the capabilities (i.e  \textit{ObjectLocalisationCapa Eq to hasComponent some ObjectTracker and hasComponent some Camera}) the agent would inherit from the capability classes and not the capability individual (i.e \textit{pepper isA ObjectLocalisationCapa} and not \textit{pepper\textunderscore{capa} isA ObjectLocalisationCapa}).
It is therefore needed to introduce a new mechanism to make available the agent's components individuals to the capability individual.
In order to solve this issue, the chained object property  \textit{hasAvailableComponent} is introduced and formalized as \textbf{\textit{isCapabilityOf o hasComponent : hasAvailableComponent}}. Thanks to this chained axiom, the capability individual is related to all of the components owned by the robot, as shown in Fig.~\ref{fig:localisationCapability}. Therefore the different capabilities enabled by the components owned by the robot individual can be inferred as classes of this capability individual. To illustrate this process, if the robot has the required components for the capability to localize objects and this latter is defined with the equivalences to \textit{hasAvailableComponent} properties instead of \textit{hasComponent}, the pepper\textunderscore{capability} individual will then be inferred as a subclass of \textit{ObjectLocalisationCapa}.
Hence, if we wanted to know what are the capabilities the robot affords, it would only require to look at which classes the capability individual inherits from.

As a reminder, other approaches required verifying with Prolog predicates which components/capabilities the robot owns to ensure that the robot has the capability to do an action. In our approach, this verification process is not needed anymore as the possession of the required components by the agent is sufficient to "enable" the capability.

\subsection{Extension to external entities}

The process described above allows generalizing to external entities.
If an agent has an object in its possession, we could represent it by a similar mechanism, allowing to infer which agent has a specific object based on the "attaching point" it is "mounted on". For example, if a robot gripper has in its claws an object, then we can logically deduce that the robot which owns the gripper has this object in its possession.
Moreover, if this specific object can provide a new capability to the agent, the process needed to describe is very similar to the one used for the components. This could be illustrated by the following example. \textit{pepper} has the capability to grasp objects and has in its hand a screwdriver which offers the \textit{ScrewingCapability}. Therefore \textit{pepper} has the capability to screw (which would be represented by the inferred \textit{pepper\textunderscore{capa} isA ScrewingCapability} relation).

\section{Discussion}

Some of the work presented here has some semantic imprecision as its purpose is rather to add functional mechanisms to the already existing ontology it has to be integrated into.
For example, we are aware that the choice made on the dependency condition over capabilities is not semantically correct, as the capability of pointing towards something with an hand depends on the capability to localize the object we want to point at, rather than being a subclass of it. Another aspect which would require more detailed work would be the \textit{hasComponent} relation which could be refined in more specific relations such as \textit{hasSoftwareComponent} or \textit{hasActuatingComponent}. This aspect is partially tackled in \cite{umbrico2020ontology} as they use the rule-inferred \textit{hasSensingPart} and \textit{hasActuatingPart} to infer the component type but this feature would need more than FOL to be inferred in our case.
This aspect is even more useful to represent the fact that the agent carries an external entity, which cannot be considered as a component (e.g the robot having a tool in its hand). The requirement for these relations must be that they should keep the same expressivity when related to the agent, and not simply be inferred as \textit{robot hasComponent tool} but rather as something like \textit{robot isCarrying tool}. 

Concerning the perception description part, we could get a more precise representation of sensors, algorithms and features of interest by using SSN and applying it towards an HRI direction, following what has been done in \cite{umbrico2020ontology}.

Moreover, as capabilities are described in this work, they currently refer to what the agent could potentially do in a neutral situation, meaning that they are not put into context. In a real environment, several key points must be tackled like verifying if a capability can effectively be used by the robot. Indeed, a robot with the capability to grab objects may not be able to effectively grab objects which are too heavy compared to its maximum payload. Another interesting feature would be having a way to describe if a capability is temporarily disabled. One could, for example, ask if when the robot is already holding an object it still has the capability to grab an object or for example to point toward something. An automatic detection of software components running on the robot (a bit similar to in \cite{roscapa}), would allow to handle software crashes and have a more real-time knowledge of what the robot can currently do. Indeed, if a software component crashes, then every inferred capability inheritance relation over this module should disappear from the robot capability individual. It could also behave similarly for a hardware component failure.


Based on what we presented in this work, we now have agent capabilities inferred based on their components. This step is one of the key requirements to model affordances as these agent-object (or agent-agent) relationships can exist only w.r.t what the agent is capable of doing. Hence, as affordances are presented in the litterature and more precisely in \cite{bessler2020foundations}, affordances related to objects could be seen as dispositions. Therefore, affordance relationships could be described by a combination of robot capabilities and object dispositions but also specific parameters required for each affordance type (e.g maximum weight liftable for the grasping affordance).






\section{Conclusion}


In this paper, we have presented an ontological pattern to describe a robot's components and capabilities. Using standard inference mechanisms, this pattern allows to infer a robot's capabilities based on its components and also other low-level capabilities. On top of easing the robot's description using ontologies, we have shown that this pattern can be applied to capabilities provided by external objects, such as tools. Indeed, when a robot grasps an object which holds a capability, it can be inferred as one of its own.


\bibliographystyle{IEEEtran}
\nocite{*}
\bibliography{biblio.bib}

\end{document}